# Decoupling anomaly discrimination and representation learning: self-supervised learning for anomaly detection on attributed graph

YanMing Hu[a,1], Chuan Chen[a,2], BoWen Deng[c,3], YuJing Lai[a,4], Hao Lin[d,5], ZiBin Zheng[b,6] and Jing Bian[a,7]

[a]*School of Computer Science and Engineering, Sun Yat-sen University, GuangZhou, China*
[b]*School of Software Engineering, Sun Yat-sen University, ZhuHai, China*
[c]*School of System Science and Engineering, Sun Yat-sen University, GuangZhou, China*
[d]*Merchants Union Consumer Finance Co., Ltd, China*

ABSTRACT

Anomaly detection on attributed graphs is a crucial topic for its practical application. Existing methods suffer from semantic mixture and imbalance issue because they mainly focus on anomaly discrimination, ignoring representation learning. It conflicts with the assortativity assumption that anomalous nodes commonly connect with normal nodes directly. Additionally, there are far fewer anomalous nodes than normal nodes, indicating a long-tailed data distribution. To address these challenges, a unique algorithm, **D**ecoupled **S**elf-supervised **L**earning for **A**nomaly **D**etection (DSLAD), is proposed in this paper. DSLAD is a self-supervised method with anomaly discrimination and representation learning decoupled for anomaly detection. DSLAD employs bilinear pooling and masked autoencoder as the anomaly discriminators. By decoupling anomaly discrimination and representation learning, a balanced feature space is constructed, in which nodes are more semantically discriminative, as well as imbalance issue can be resolved. Experiments conducted on various six benchmark datasets reveal the effectiveness of DSLAD.

## 1. Introduction

To display the intricate and interconnected data, attributed graphs are frequently employed. Recently, anomaly detection on attributed graphs has attracted lots of interest, which seeks to identify some minority patterns (such as nodes, and edges.) that deviate from the majority tremendously on the graph [17]. Anomaly detection on the attributed graph can be deployed in many real-world scenarios, such as spotting fraud in transaction networks, spotting incorrect citation relations among academic papers, and spotting users who deliver spam in postal transportation networks.

However, anomaly detection on attributed graphs is quite a challenging task that primarily faces three challenges. First, it is a heavy cost to obtain enough labels for anomalous nodes. Therefore, supervised models are not applicable for anomaly detection, as evidenced by the fact that ground-truth labels and the class of anomalies are always unknown [3]. Second, anomalous nodes' neighbors are commonly normal nodes. GNN-based algorithms largely rely on aggregating messages from neighbors [32, 8, 1, 25]. As a consequence, anomalous nodes are buried by messages of normal nodes, leading to the mixture in semantic space. Third, the number of anomalous nodes is far less than that of normal nodes. Traditional deep learning algorithms suffer from imbalance issue [14, 6, 36, 29] that the majority dominates the embedding training and the minority is often mistakenly identified as the majority. Therefore, it is urgent to propose an effective self-supervised algorithm to address the above three challenges for anomaly detection.

Several methods for anomaly detection on attributed graphs have been proposed. These methods have achieved great success in anomaly discrimination, but they still have some drawbacks. The shallow methods, such as AMEN [20], are limited by the capacity for expressiveness. The node-classification-targeted methods, such as DOMINANT [3], simply combining existing models for node classification and an anomaly discriminator, are not directly designed for anomaly detection. The anomaly-detection-targeted methods, such as CoLA [15] optimize the model directly for anomaly detection, but they mainly revolve around anomaly discrimination, paying insufficient attention to representation learning, which leads to semantic mixture and imbalance issue.

To overcome the aforementioned challenges, in this paper, we propose a novel method DSLAD for anomaly detection. In DSLAD, both contrastive learning and generative learning are adopted to discriminate anomaly. Especially, DSLAD contrasts node-subgraph pairs and measures reconstruction errors to calculate anomaly scores. The anomaly score is further categorized into context anomaly score and reconstruction anomaly score, deployed with bilinear pooling and masked autoencoder respectively as anomaly discriminator. Considering semantic mixture and imbalance issue, we introduce contrastive representation learning and decouple it with anomaly discrimination. Through decoupling anomaly discrimination and contrastive representation learning, DSLAD maps nodes into a balanced semantic space with a little semantic mixture.

∗∗Corresponding author
huym27@mail2.sysu.edu.cn (Y. Hu); chenchuan@mail.sysu.edu.cn (C. Chen); bowen.deng20@gmail.com (B. Deng); laiyj23@mail2.sysu.edu.cn (Y. Lai); linhao@mucfc.com (H. Lin); zhzibin@mail.sysu.edu.cn (Z. Zheng); mcsbj@mail.sysu.edu.cn (J. Bian)
ORCID(s):

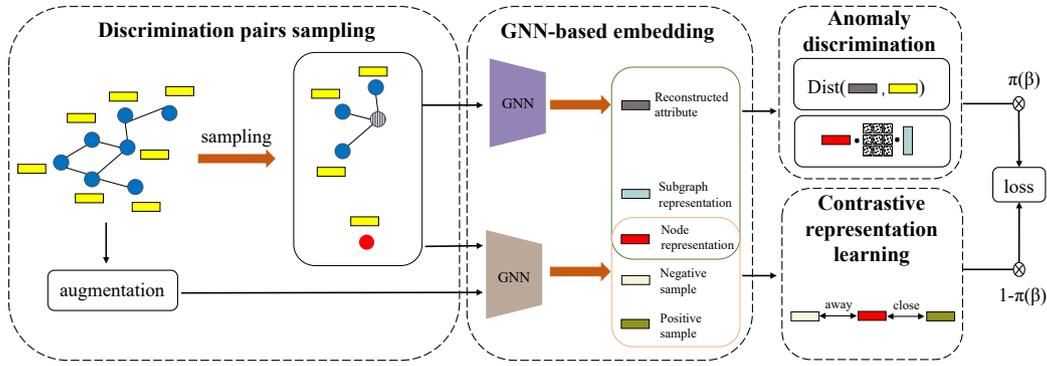

**Figure 1:** Framework of DSLAD. There are four components in DSLAD: Discrimination pairs sampling, GNN-based embedding, Anomaly discrimination and Contrastive representation learning. DSLAD firstly selects a set of target nodes and samples neiboring subgraphs of them. Next, the nodes and the sampled subgraphs are embedded into low dimension vectors by GNN for anomaly detection and contrastive representation learning. Finally, the discriminators measures the distance between node-subgraph pairs to discriminate anomaly while target nodes are pulled close to positive samples and pushed away from negative samples (when training only). Especially, anomaly discrimination and representation learning are decoupled.

The contributions of this work can be summarized as follows:

- We integrate contrastive representation learning into the anomaly detection model, which makes nodes more semantically distinguishable and vastly benefits anomaly discrimination.
- We decouple contrastive representation learning and anomaly discrimination, ulteriorly resolving semantic mixture and imbalance issue in anomaly detection.
- We conduct a series of experiments on six datasets and the results demonstrate the superiority of DSLAD over the existing models.

## 2. Related work
### 2.1. Graph Neural Networks

Recently, graph representation learning has achieved considerable success with GNNs. The core idea of GNNs is aggregating messages from neighbors to update node representations, which is based on the assortativity assumption. GNNs can be divided into two categories: spectral-based methods and spatial-based methods. The former category includes GCN [8], passing message by first-order approximation of Chebyshev filter. The latter category includes GAT [25] and GraphSAGE [4]. GAT utilizing attention mechanism, assigns weight to each edge when aggregating messages. GraphSAGE proposes inductive representation learning manner to cope with tasks on large-scale graphs.

Apart from above fundamental GNN models, many advanced GNN models are also proposed to learn the graph representations better. To avoid the sparsity issue and filter the noise information, [12] proposes a framework preserving low-order proximities, mesoscopic community structure information and attribute information for network embedding. MTSN [16], a dynamic graph neural network, captures local high-order graph signals, learns attribute information based on motifs, and preserve timing information by temporal shifting. To alleviate the oversmooth issue, NAIE [2] adopts an adaptive strategy to smooth attribute information and topology information, and develop an autoencoder to enhance the embedding capacity.

### 2.2. Graph self-supervised learning

Self-supervised graph learning, a new learning paradigm that trains models without labels, has been widely used in computer vision [9] and natural language processing [10]. Self-supervised learning in graph can be categorized into : graph contrastive learning (e.g. SimGRACE [30]), graph generative learning (e.g. Graph Completion [35]) and graph predictive learning (e.g. CDRS [39]). Without augmentation, SimGRACE, uses a formal encoder and a perturbation encoder to embed the graph, then pulls close the same semantic while pushing away the different semantics among the two hidden spaces. Graph Completion removes features of the target node, and then reconstructs it from the unmasked neighboring nodes. CDRS makes a pseudo node classification task collaborated with the clustering task to enhance representation learning.

### 2.3. Anomaly detection on attributed graph

Anomaly detection on attributed graph works for identifying patterns that notably diverge from the majority. Many methods have been proposed for anomaly detection, including the shallow methods and the deep methods. The shallow methods include [20], Radar [11], and ANOMALOUS [19]. AMEN measures the correlation of features between the target node and its ego-networks to detect anomaly. Radar analyzes the residuals of attribute information and its coherence with graph information to detect anomaly. ANOMALOUS integrates CUR decomposition and residual analysis to detect anomaly. The shallow methods are limited by their expressiveness ability in graph embedding. The deep methods can be further divided into two classes. The first class deep methods include DOMINANT [3] and DGIAD [26, 15]. DOMINANT reconstructs the adjacency matrix and the attribute matrix, then distinguish anomaly through reconstruction error. DGI contrasts node and graph for embedding. Deployed with a trained discriminator, DGI can be used for anomaly detection and we rename this method as DGIAD in this paper. The first class deep methods, node classification models merely equipped with anomaly discriminator, are not devised for anomaly detection and still don't show satisfying performance. The second class deep methods include CoLA [15], SL-GAD [37], and ANEMONE [7]. CoLA, SL-GAD and ANEMONE contrast nodes and subgraphs to discriminate anomaly. The second class deep methods optimize model toward anomaly detection, but they neglect representation learning to overcome semantic mixture and imbalance issue.

## 3. Problem definition

In this section, the problem definition of anomaly detection on attributed graph will be introduced. Given an attributed graph $\mathcal{G} = (\mathcal{V}, \mathbf{X}, \mathbf{A})$, the target of anomaly detection is to learn a mapping mechanism $\mathcal{F}(\cdot)$ to calculate the anomaly score $\mathbf{s}_i, i \in \mathcal{V}$ for nodes in $\mathcal{G}$. The anomaly score $\mathbf{s}_i$ describes the abnormal degree of the node $i$. It is easy to detect anomaly, if the mapping mechanism $\mathcal{F}(\cdot)$ is well designed and outputs accurate anomaly scores. For the convenience of reading this paper, all important notations are explained in Table 1.

## 4. Method

In this section, a thorough introduction to DSLAD will be given. As shown in Figure 1, DSLAD consists of four modules, discrimination pair sampling, GNN-based embedding, anomaly discrimination and contrastive representation learning. On attributed graph, contrastive learning at the node-subgraph level is powerful for graph representation learning [31, 13]. Tt has been discovered that detecting anomalies at the node-subgraph level is effective [15]. To detect the anomaly, we sample discrimination pairs at node-subgraph level. The target nodes and their sampled subgraphs are then embedded into low dimension vectors via GNN. Next, the embedding vectors of target nodes and their sampled subgraphs are fed into anomaly discrimination and contrastive representation learning. By contrastive representation learning, the semantic mixture and imbalance issue can be lightened, and decoupling anomaly discrimination and contrastive representation learning can further alleviate it.

**Table 1**
Statements of important notations

| notations | statements |
|---|---|
| $\mathcal{G} = (\mathcal{V}, \mathbf{X}, \mathbf{A})$ | An attributed graph |
| $v_i$ | The i-th node in $\mathcal{G}$ |
| $\mathcal{G}_i$ | The subgraph originated from $v_i$ |
| $K$ | The number of nodes in $\mathcal{G}_i$ |
| $\mathbf{X} \in \mathbb{R}^{N \times d(0)}$ | The attribute matrix of $\mathcal{G}$ |
| $\mathbf{A} \in \mathbb{R}^{N \times N}$ | The adjacency matrix of $\mathcal{G}$ |
| $\mathbf{X}_{\{i\}} \in \mathbb{R}^{K \times d(0)}$ | The attribute matrix of $\mathcal{G}_i$ |
| $\mathbf{A}_{\{i\}} \in \mathbb{R}^{K \times K}$ | The adjacency matrix of $\mathcal{G}_i$ |
| $\mathbf{h}_{\{i\}}^{(l)} \in \mathbb{R}^{1 \times d(l)}$ | The embedding of $v_i$ in the $l$-th layer |
| $\mathbf{W}^{(l)} \in \mathbb{R}^{d(l-1) \times d(l)}$ | The weight matrix in the $l$-th layer |
| $\mathbf{H}_{\{i\}}^{(l)} \in \mathbb{R}^{N \times d(l)}$ | The hidden matrix in the $l$-th layer of $\mathcal{G}_i$ |
| $\mathbf{Z}^i \in \mathbb{R}^{K \times d}$ | The context representation matrix of $\mathcal{G}_i$ |
| $\mathbf{U}_{\{i\}} \in \mathbb{R}^{N \times d(0)}$ | The reconstructed attribute matrix of $\mathcal{G}_i$ |
| $\mathbf{g}_i \in \mathbb{R}^{1 \times d}$ | The subgraph-level representation of $\mathcal{G}_i$ |
| $\mathbf{e}_i \in \mathbb{R}^{1 \times d}$ | The node-level representation of $v_i$ |
| $x_i \in \mathbb{R}^{1 \times d(0)}$ | The attribute of $v_i$ |
| $\mathbf{W}_d \in \mathbb{R}^{d \times d}$ | The weight matrix of bilinear pooling |
| $\mathbf{s}_i^{con(-)}$ | The negative context anomaly score of $v_i$ |
| $\mathbf{s}_i^{con(+)}$ | The positive context anomaly score of $v_i$ |
| $\mathbf{s}_i^{rec}$ | The reconstruction anomaly score of $v_i$ |
| $\mathbf{s}_i$ | The anomaly score of $v_i$ |

### 4.1. Discrimination pair sampling.

The key to anomaly detection is finding the patterns significantly different from the majority. Therefore, discrimination pairs are crucial to this task. Graph objects can be categorized into edge, node, subgraphs and graph. Any two of them, excluding edge, can be selected to constitute discrimination pairs. We sample discrimination pairs at node-subgraph level. The procedure is as follows:

- **Target node selection.** A set of nodes are randomly selected from the input graph every epoch without replacement so that each node has the same chance of being chosen.

- **Subgraph sampling.** For every selected target node, a neighboring subgraph is sampled via random walks with restart (RWR) [24] as augmentation, avoiding introducing extra anomalies. Other sampling methods also can be considered. The size of the neighboring subgraph is fixed to $K$, which determines the scope of the target node for matching.

DSLAD- **Attribute mask.** The attributes of the target node are masked with zero vectors in the sampled subgraph, making it more difficult to identify the information of the target node in the subgraph. This mechanism will improve the ability of anomaly detection [15, 37].

Target nodes and neighboring subgraphs are combined as discrimination pairs for anomaly discrimination. A positive pair includes a node and a subgraph sampled from it, while a negative pair includes a node and a subgraph sampled from other nodes.

### 4.2. GNN-based embedding

For anomaly discrimination and contrastive representation learning, obtained target nodes and their neighboring subgraphs are mapped into low-dimensional embedding space by GNNs.

We apply a GCN encoder and a GCN autoencoder to embed the graph and reconstruct the attribute matrix, respectively.

Target node $v_i$ is embedded as a graph with only one node. GNN propagation formula can thus be simplified to MLP:

$$\mathbf{h}_i^{(l+1)} = \sigma(\mathbf{h}_i^{(l)}\mathbf{W}^{(l)}). \quad (1)$$

And $\mathbf{e}_i \in \mathbb{R}^d$ is used to denote the output of GCN encoder, which is the node level representation vector of $v_i$.

On the $K$-nodes subgraph $\mathcal{G}_i$ sampled from node $v_i$, the adjacency matrix is denoted by $\mathbf{A}_{\{i\}} \in \mathbb{R}^{K \times K}$ and the attribute matrix is denoted by $\mathbf{X}_{\{i\}} \in \mathbb{R}^{K \times d(0)}$. Then, the GNN operator is applied to:

$$\mathbf{H}_{\{i\}}^{(l+1)} = \sigma(\widetilde{\mathbf{D}}^{-\frac{1}{2}}\widetilde{\mathbf{A}}_{\{i\}}\widetilde{\mathbf{D}}^{-\frac{1}{2}}\mathbf{H}_{\{i\}}^{(l)}\mathbf{W}^{(l)}), \quad (2)$$

where $\widetilde{\mathbf{A}}_{\{i\}} = \mathbf{A}_{\{i\}} + \mathbf{I}_K$, and $\mathbf{H}_{\{i\}}^{(0)} = \mathbf{X}_{\{i\}}$.

The output of the GCN encoder is denoted by $\mathbf{Z}^i \in \mathbb{R}^{K \times d}$, which is the context representation matrix of subgraph $\mathcal{G}_i$. And the output of the GCN autoencoder on $\mathcal{G}_i$ is denoted by $\mathbf{U}_{\{i\}} \in \mathbb{R}^{K \times d(0)}$, which is the reconstructed attribute matrix of $\mathcal{G}_i$.

The readout module summarizes $\mathbf{Z}^i$ into its subgraph-level representation $\mathbf{g}_i \in \mathbb{R}^d$. We take average pooling as the readout module. The subgraph-level representation can then be formulated as:

$$\mathbf{g}_i = readout(\mathbf{Z}^i) = \frac{1}{K-1}\sum_{j=0, j\neq c_i}^{K}\mathbf{Z}^i[j,:], \quad (3)$$

where $c_i$ is the index of $v_i$ in neighboring subgraph $\mathcal{G}_i$.

### 4.3. Anomaly discrimination

In this subsection, We will describe how DSLAD discriminates anomaly. Context anomaly and reconstruction anomaly are two subtypes of anomaly discrimination. Here is more information on them in depth:

#### 4.3.1. Context anomaly

Anomalies differ from the other majority significantly. The anomalous nodes are supposed to be far away from normal nodes in the embedding space. To assess how a discrimination pair matches, we take bilinear pooling as the discriminator.

Given a node $v_i$ and the relevant neighboring subgraph $\mathcal{G}_i$, the context anomaly score of this discrimination pair can be calculated as:

$$\mathbf{s}_i^{con} = disc(\mathbf{g}_i, \mathbf{e}_i) = \phi(\mathbf{g}_i\mathbf{W}_d\mathbf{e}_i^T), \quad (4)$$

where $\mathbf{W}_d \in \mathbb{R}^{d \times d}$ is a learnable weight matrix, and $\phi(\cdot)$ is non-linear and non-negative activation function. Here we use Sigmoid as the activation function.

We use graph contrastive learning at the node-subgraph level, taking both positive and negative discrimination pairs into account. For target node $v_i$, we take $P$ positive discrimination pairs and $Q$ negative discrimination pairs to compute context anomaly score. The positive score $\mathbf{s}_i^{con(-)}$ and the negative score $\mathbf{s}_i^{con(+)}$ are formulated as:

$$\mathbf{s}_i^{con(-)} = \frac{1}{Q}\phi(\sum_{\mathbf{g}_j \in \{\mathbf{g}^{(-)}\}}\mathbf{g}_j\mathbf{W}_d\mathbf{e}_i^T), \quad (5)$$

$$\mathbf{s}_i^{con(+)} = \frac{1}{P}\phi(\sum_{\mathbf{g}_j \in \{\mathbf{g}^{(+)}\}}\mathbf{g}_j\mathbf{W}_d\mathbf{e}_i^T), \quad (6)$$

where $\{\mathbf{g}^{(-)}\} \in \mathbb{R}^d$ and $\{\mathbf{g}^{(+)}\} \in \mathbb{R}^d$ denote the positive neighboring subgraph set and the negative neighboring subgraph set for the target node $v_i$, respectively. For simplicity, we set $P = Q = 1$.

In this part, our optimization goal is maximizing the agreement with the context anomaly score and the ground-truth label (label 1 for positive pairs and 0 for negative pairs). The loss function of context anomaly score can be formulated as:

$$L_{con} = -\frac{1}{2|\mathcal{V}|}\sum_{i \in \mathcal{V}}(log(s_i^{con(+)}) + log(1 - s_i^{con(-)})) \quad (7)$$

#### 4.3.2. Reconstruction anomaly

Inspired by [37, 35], we introduce the reconstruction error as a supplementary mechanism to anomaly discrimination. For target node $v_i$, we have removed its attributes on neighboring subgraph $\mathcal{G}_i$. DSLAD tries to reconstruct the attributes of the target node $v_i$, from the other nodes on $\mathcal{G}_i$. $l_2$-norm is adopted to measure the distance between original information and reconstructed information quantitatively.

The index of $v_i$ in the neighboring subgraph $\mathcal{G}_i$ is $c_i$. The reconstructed attribute vector of $v_i$ in neighboring subgraph $\mathcal{G}_i$ is denoted by $\mathbf{U}_{\{i\}}[c_i,:]$.

In order to train the masked autoencoder for reconstruction anomaly, we adopt MSE as the loss function for this portion. It can be written as:

$$L_{rec} = -\frac{1}{|\mathcal{V}|}\sum_{i \in \mathcal{V}}||\mathbf{U}_{\{i\}}[c_i,:] - x_i||^2, \quad (8)$$

where $x_i \in \mathbb{R}^{d(0)}$ is the original attribute vector of node $v_i$.

## 4.4. Contrastive representation learning

In the context anomaly module, the loss function in equation (7) mainly focuses on anomaly discrimination and pays less attention to representation learning. Additionally, it assumes that all nodes contribute equally, leading to semantic mixture and imbalance issue. To impede the normal nodes from dominating the representation learning, we implement the contrastive representation learning module and set the number of positive samples and negative samples equal.

For target node $v_i$, we select neighboring subgraph $\mathcal{G}_j$ ($j \neq i$) sampled from node $v_j$ as the negative sample, in consistency with context anomaly module. As for the positive sample, the neighboring subgraph $\mathcal{G}_i$, augmented by the strategy *local_aug*, is selectable. Another augmentation strategy, denoted by *global_aug*, embeds the whole graph without the mask. Target node augmented by *global_aug* can also be the positive sample. $\mathbf{e}_i^-$ denotes representation vector of negative sample, and $\mathbf{e}_i^- = \mathbf{g}_j, j \neq i$. $\mathbf{e}_i^+$ denotes representation vector of positive sample, and

$$\mathbf{e}_i^+ = \begin{cases} \mathbf{g}_i & local\_aug \\ f(\mathbf{X}, \mathbf{A})[i, :] & global\_aug \end{cases}, \quad (9)$$

where $\mathbf{g}_i \in \mathbb{R}^d$ is computed by equation (3) and $f$ denotes GCN encoders in context anomaly module.

Both the number of positive samples and negative samples are set to 1 for simplicity and fairness. We adopt infoNCE [18] as the loss function of contrastive representation learning:

$$L_{CL} = \frac{1}{|\mathcal{V}|} \sum_{i \in \mathcal{V}} -\log \frac{\exp^{(\mathbf{e}_i \cdot \mathbf{e}_i^+/\tau)}}{\exp^{(\mathbf{e}_i \cdot \mathbf{e}_i^+/\tau)} + \exp^{(\mathbf{e}_i \cdot \mathbf{e}_i^-/\tau)}}, \quad (10)$$

where $\tau$ is a temperature parameter greater than 0.

## 4.5. Decoupling

In this subsection, we explain why and how we decouple anomaly discrimination and contrastive representation learning. When behaviors and label semantics are excessively inconsistent in anomaly detection tasks, [28] has shown that training graph representation learning and anomaly discrimination jointly may lead to performance degradation. Moreover, the problem of class imbalance can also be resolved significantly by decoupling representation learning and anomaly discrimination. At the beginning of training, discriminators are prone to predict arbitrarily, producing erroneous results while contrastive representation learning forms a balanced semantic space [33, 5, 34]. During training, discriminators makes increasingly accurate predictions while performance gained by contrastive representation learning decays [38, 27]. Gradually shifting to anomaly discrimination from contrastive learning enhances the effectiveness. Based on the above analysis, instead of jointly training by classification loss, we decouple anomaly discrimination and contrastive representation learning and give them dynamic weights.

Let $\beta$ denotes the ratio of current epoch to the number of training epochs, whose value indicates the training process.

$\pi(\beta)$ is the factor balancing the anomaly discrimination loss and the contrastive representation learning loss, where $\pi(\cdot)$ is a mapping function. The final loss function can be written as:

$$L = \pi(\beta)(\alpha L_{con} + (1-\alpha)L_{rec}) + \lambda(1-\pi(\beta))L_{CL}, \quad (11)$$

where $\alpha$ and $\lambda$ are the hyperparameter that control the role of different anomaly scores, and scale contrastive representation learning, respectively. And $\pi(\beta)$ increases with $\beta$.

## 4.6. Anomaly score calculation

We could calculate the final anomaly score for each node after training.

For node $v_i$, context anomaly score $\mathbf{s}_i^{con}$ and reconstruction anomaly score $\mathbf{s}_i^{rec}$ can be inferred as follows:

$$\mathbf{s}_i^{con} = \mathbf{s}_i^{con(-)} - \mathbf{s}_i^{con(+)}, \quad (12)$$

where $\mathbf{s}_i^{con(-)}$ and $\mathbf{s}_i^{con(+)}$ are calculated by equation (5) and (6) respectively.

$$\mathbf{s}_i^{rec} = ||\mathbf{U}_{\{i\}}[c_i, :] - x_i||_2^2, \quad (13)$$

where the index of $v_i$ in the neighboring subgraph $\mathcal{G}_i$ is $c_i$, the reconstructed attribute vector of $v_i$ in neighboring subgraph $\mathcal{G}_i$ is $\mathbf{U}_{\{i\}}[c_i, :]$, and $x_i$ is the original attribute vector of node $v_i$.

By MinMaxScalar, we transform the context anomaly score $\mathbf{S}^{con}$ to [0,1] for standardization:

$$\mathbf{s}_i^{con} = \frac{\mathbf{s}_i^{con} - \mathbf{s}_{min}^{con}}{\mathbf{s}_{max}^{con} - \mathbf{s}_{min}^{con}}, \quad (14)$$

where $\mathbf{s}_{min}^{con}$ and $\mathbf{s}_{max}^{con}$ are the min and the max of context anomaly scores, respectively. Similarly, reconstruction anomaly score $\mathbf{S}^{rec}$ is also transformed to [0,1] by MinMaxScalar:

$$\mathbf{s}_i^{rec} = \frac{\mathbf{s}_i^{rec} - \mathbf{s}_{min}^{rec}}{\mathbf{s}_{max}^{rec} - \mathbf{s}_{min}^{rec}}, \quad (15)$$

where $\mathbf{s}_{min}^{rec}$ and $\mathbf{s}_{max}^{rec}$ are the min and the max of reconstruction anomaly scores, respectively.

Combining transformed context anomaly score and reconstruction anomaly score, we can get the final anomaly score $\mathbf{s}_i$ of node $v_i$:

$$\mathbf{s}_i = \alpha \mathbf{s}_i^{con} + (1-\alpha)\mathbf{s}_i^{rec}. \quad (16)$$

Neighboring subgraph is sampled stochasticly. To reduce the sampling variance, we take the averaging anomaly score over $R$ times as the final anomaly score.

## 5. Experiments

In this part, a succession of experiments are carried out on six real-world datasets to examine the effectiveness of our model.

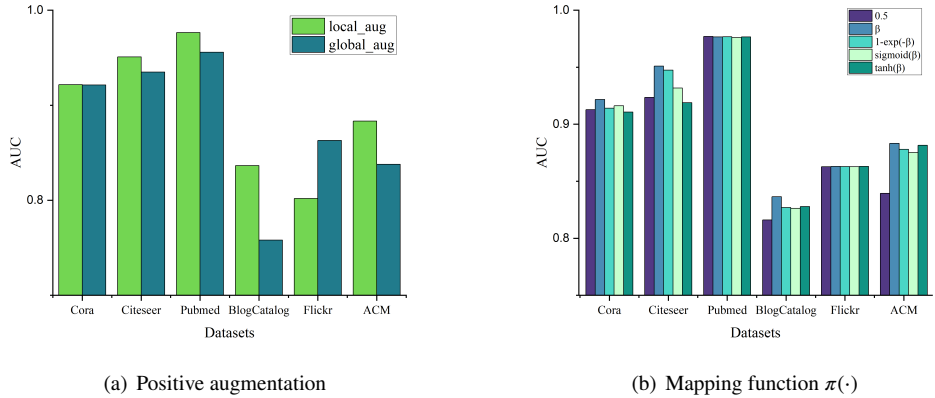

(a) Positive augmentation  (b) Mapping function $\pi(\cdot)$

**Figure 2**: Performance comparison between different positive augmentation strategies and mapping function $\pi(\cdot)$

**Table 2**
The statistics of datasets.

| Dataset | Anomalies | nodes | Features | Edges |
|---|---|---|---|---|
| Cora | 150 | 2,708 | 1,433 | 5,429 |
| Citeseer | 150 | 3,327 | 3,703 | 4,732 |
| Pubmed | 600 | 19,717 | 500 | 44,338 |
| ACM | 600 | 16,484 | 8,337 | 71,980 |
| BlogCatalog | 300 | 5,196 | 8,189 | 171,743 |
| Flickr | 450 | 7,575 | 12,407 | 239,739 |

### 5.1. Datasets

Six frequently used real-world datasets for anomaly identification, including four citation network datasets and two social network datasets, are applied to evaluate our model.

The following is a brief overview of the six datasets:

- **Citation network datasets.** Cora, Citeseer, Pubmed [21] and ACM [22] are four public citation network datasets, composed of scientific publications. In the four citation networks, the published papers are transformed into nodes while edges represent the citation relationships between papers. And the description text of papers can be transformed into nodes features.

- **Social network datasets.** BlogCatalog and Flickr [23] are acquired from the websites for sharing blogs and images, respectively. In the two datasets, each user is represented by a node, and links among nodes illustrate the relationships between corresponding users. Users often describe themselves with personalized information, such as posting blogs and public photos. Features can be extracted from such information.

Considering that there are no ground-truth anomaly labels in above six real-world datasets, injecting synthetic anomaly nodes into datasets to simulate real anomalies is used widely. We follow the perturbation processing in [15, 3] to inject anomalies with both attribute anomalies and structure anomalies into the six datasets. For the attribute anomaly injection, we select $M_a$ nodes and replace their features with stochasticly selected remote nodes. For the structure anomaly injection, we pick up $M_a$ nodes and divide them into $M_c$ clusters averagely. Nodes within the same cluster are connected with each other. The statistics of these contaminated datasets are depicted in Table 2.

### 5.2. Baselines

We choose some of the state-of-the-art methods as baselines to compare with our proposed DSLAD on the above six real-world datasets. These methods are divided into three categories:

(1) The shallow method: The shallow method detects anomalies without deep learning. We pick up the following three models for comparison:

- **AMEN** [20] compares the correlation of features of the target nodes and theri ego-networks to identify the nodes with low scores as anomalies.

- **Radar** [11] analyzes the residuals of attribute information and its coherence with graph information to detect the abnormal nodes as anomalies.

- **ANOMALOUS** [19] utilizes CUR decomposition and residual analysis to distinguish the irregular nodes as anomalies.

(2) The node-classification-targeted method: The node-classification-targeted method simply expand the node classification model with an anomaly detection module to detect anomaly. We choose the following two models for comparison:

- **DOMINANT** [3] learns node embeddings by autoencoders and take the reconstruction errors as the anomaly scores.

**Table 3**
Comparison experiment results of anomaly detection by AUC metric on six benchmark datasets. The best performance and the second-best performance methods are marked by bold and underlined fonts respectively. P-value=0.0331 (popmean=mean+std), and the std can be seen in Table 4.

| Methods | Cora | Citeseer | Pubmed | BlogCatalog | Flickr | ACM |
|---|---|---|---|---|---|---|
| AMEN | 0.6266 | 0.6154 | 0.7713 | 0.6392 | 0.6573 | 0.5626 |
| Radar | 0.6587 | 0.6709 | 0.6233 | 0.7401 | 0.7399 | 0.7247 |
| ANOMALOUS | 0.5770 | 0.6307 | 0.7316 | 0.7237 | 0.7434 | 0.7038 |
| DOMINANT | 0.8155 | 0.8251 | 0.8081 | 0.7468 | 0.7442 | 0.7601 |
| DGIAD | 0.7511 | 0.8293 | 0.6962 | 0.5827 | 0.6237 | 0.6240 |
| CoLA | 0.8799 | 0.8968 | 0.9512 | 0.7854 | 0.7513 | 0.8237 |
| SL-GAD | <u>0.9130</u> | 0.9136 | <u>0.9672</u> | <u>0.8184</u> | <u>0.7966</u> | 0.8538 |
| ANEMONE | 0.9057 | <u>0.9189</u> | 0.9548 | 0.8067 | 0.7637 | <u>0.8709</u> |
| Ours | **0.9196** | **0.9481** | **0.9772** | **0.8275** | **0.8631** | **0.8809** |

**Table 4**
Standard Deviation and P-value on six datasets

| | mean | std | P-value(mean+std) |
|---|---|---|---|
| Cora | 0.9196 | 0.0028 | |
| Citeseer | 0.9481 | 0.0027 | |
| Pubmed | 0.9772 | 0.0004 | 0.0331 |
| ACM | 0.8809 | 0.0022 | |
| BlogCatalog | 0.8275 | 0.0041 | |
| Flickr | 0.8631 | 0.0017 | |

- **DGIAD** [26, 15] uses DGI to learn node embeddings and take the bilinear pooling to compute the anomaly scores.

(3) The anomaly-detection-targeted method: The anomaly-detection-targeted method is designed to detect anomaly directly, even without considering node classification. We select the following three models for comparison:

- **CoLA** [15] learns node embedding by GCN and contrasts nodes and subgraphs to discriminate anomaly.
- **ANEMONE** [7] expands **CoLA** with the patch-level contrast.
- **SL-GAD** [37] expands **CoLA** with the reconstruction error.

### 5.3. Evaluation metrics

We utilize ROC-AUC, a widely used metric for anomaly detection, to quantify the performance of DSLAD and the baselines. The ROC curve is depicted by the true positive rate (y-variable) and the false positive rate (x-variable). AUC is the area enclosed by the ROC curve the x-axis. AUC always falls between 0 and 1. The better performance is indicated by the higher AUC.

### 5.4. Experiments setting

We set neighboring subgraph size $k \in Z^+$ in [2,10]. Layers of GNN encoders and GNN decoders are set as 3 on Flickr, and 1 on the other five datasets. The hidden dimension $d$ is set as 64 while test rounds $R = 256$. The batch size is 300. We choose $\pi(\beta) = \beta$ and select $\lambda$ from $\{0.5, 1, 1.5, 2, 2.5, 3\}$. Our proposed model is implemented in server with Ubuntu 20.04.1 LTS, Pytorch 1.10, dgl0.4.3post2, Intel(R) Xeon(R) Gold 6132 CPU @ 2.60GHz, and GeForce RTX 2080 Ti. We execute DSLAD over 8 times to measure the effectiveness statistically.

### 5.5. Comparison Results

To verify the effectiveness of our model in anomaly detection task, we conducted comparison experiments for all baselines and DSLAD with AUC metric on six benchmark datasets and results are shown in Table 3. Based on the results, we can make the following observations:

- Compared with the most advanced baselines, our method outperforms baselines on all benchmark datasets with a large margin, improving 0.72% at least, 8.35% at most and 2.59% on average. It reveals the effectiveness of our method.
- The shallow methods AMEN, Radar, and ANOMALOUS perform worse than other baselines because of the limitation of expressiveness capacity.
- The node-classification-targeted methods DOMINANT and DGIAD perform better than shallow methods. DOMINANT reconstructs attribute matrix and adjacency matrix, not directly targeting to detect anomalies. DGIAD contrasts the nodes and the whole graph, utilizing very little local information.
- The anomaly-detection-targeted methods CoLA, ANEMONE and SL-GAD make a step further. However, they mainly concentrate on training anomaly discriminator, still constrained by semantic mixture and imbalance issue.

**Table 5**
Ablation studies on six benchmark datasets. How each module would effect the whole model is explored. Variants DSLAD w/o cl, DSLAD w/o con and DSLAD w/o rec are generated by removing contrastive representation learning and set $\pi(\cdot)=1$, removing context score and removing reconstruction score, respectively.

| Variants | Cora | Citeseer | Pubmed | BlogCatalog | Flickr | ACM |
|---|---|---|---|---|---|---|
| DSLAD | **0.9196** | **0.9481** | **0.9772** | **0.8275** | **0.8631** | **0.8809** |
| DSLAD w/o cl | 0.8948 | 0.9357 | 0.9726 | 0.8163 | 0.7883 | 0.8232 |
| DSLAD w/o con | 0.8275 | 0.8036 | 0.8050 | 0.7474 | 0.744 | 0.7463 |
| DSLAD w/o rec | 0.9060 | 0.9023 | 0.9545 | 0.7982 | 0.6954 | 0.8565 |

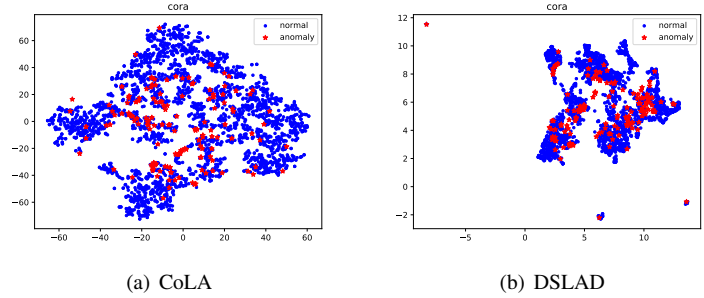

(a) CoLA  (b) DSLAD

**Figure 3:** The visualization of CoLA embeddings and DSLAD context embeddings.

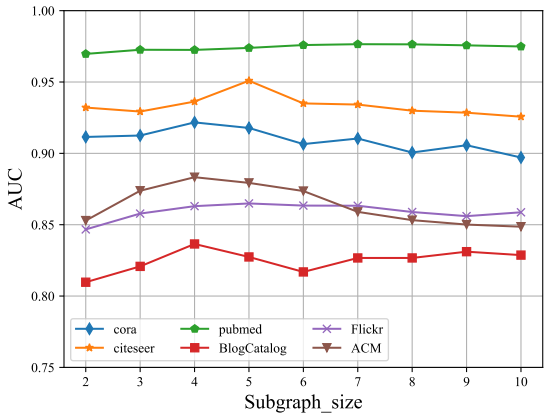

**Figure 4:** Performance comparison using different subgraph size on the six benchmark datasets

### 5.6. Augmentation strategy

In this subsection, we assess the effectiveness of the augmentation strategy to our method. As illustrated in Figure 2 (a), our model is not sensitive to the augmentation strategy on Cora, and performs better with *local_aug* than with *global_aug* on the other datasets except for Flickr.

The main reason may be that Flickr has the most complex attribute information, which is more important than structure information. And on the other five dataset, structure information has a greater impact on the other five datasets than attribute information.

Based on the above observations, we choose *global_aug* as augmentation strategy on Flickr, and *local_aug* as augmentation strategy on the rest five datasets.

### 5.7. $\pi(\beta)$ strategy

In this subsection, we investigate how different mapping functions $\pi(\beta)$ would effect our model. Constant e.g. 0.5, linear growth e.g. $\beta$, and activation function e.g. $1 - exp(-\beta)$, $sigmoid(\beta)$, and $tanh(\beta)$ are taken into consideration. As demonstrated in Figure 2 (b), when setting $\pi(\beta) = \beta$, the best results are acquired, although our model is not sensitive to $\pi(\beta)$ on Pubmed and Flickr. Linear $\pi(\beta)$ also has the best generalization.

### 5.8. Ablation studies

In this subsection, we conduct ablation studies for better understanding the effectiveness of each components in our method. Here, three variants are defined as:

- **DSLAD w/o cl**: remove contrastive representation learning and set $\pi(\beta) = 1$.
- **DSLAD w/o con**: remove context score.
- **DSLAD w/o rec**: remove reconstruction score.

As shown in Table 5, DSLAD outperforms other variants, indicating that all components play an important role in our method and they could make mutual promotion. We demonstrate the visualization of embedding on Cora in Figure 3. In Figure 3, we can find that both the normal nodes and the anomalous nodes are more dense learned by DSLAD than CoLA, and DSLAD even works for node classification. Removing contrastive representation learning causes remarkable performance degradation. Obviously, contrastive representation learning promotes anomaly discrimination a lot by reducing class imbalance and lightening semantic mixture. Additionally, removing either context or reconstruction scores performance would result in performance degradation with the former being more noticeable. This demonstrates that context score and reconstruction score complement each other, while context scores are more effective for anomaly detection.

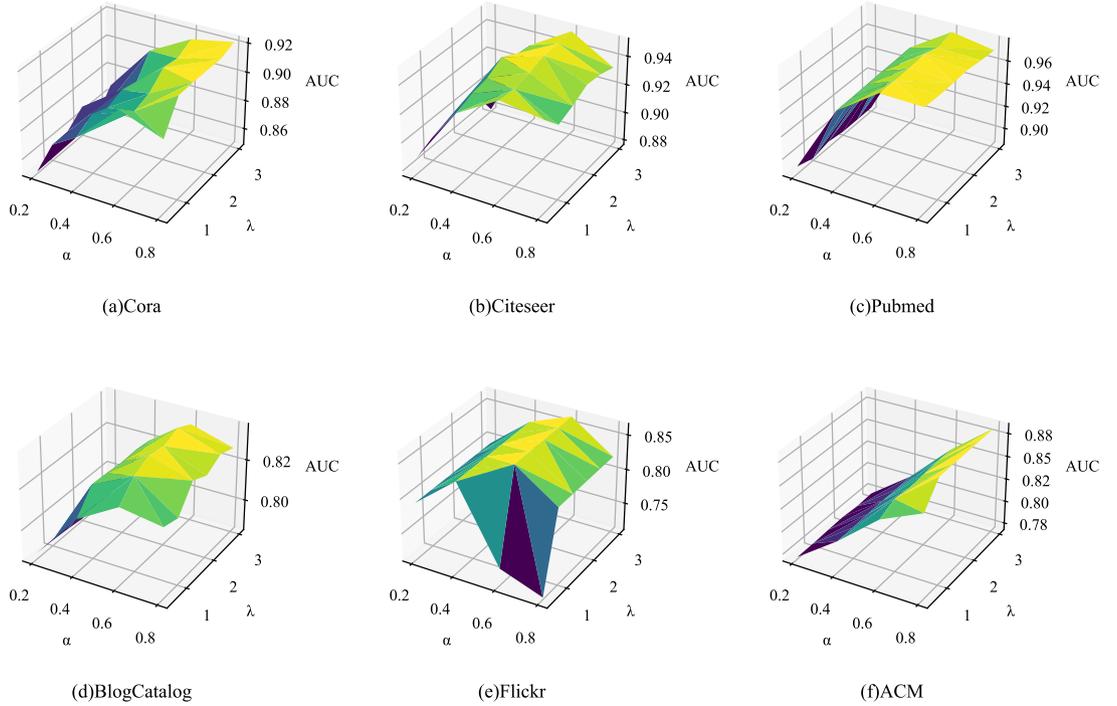

**Figure 5**: Parameter sensitivity studies for hyperparameter $\alpha$ and $\lambda$. The values of AUC are correlated with colors by viridis.

### 5.9. Parameter sensitivity

In this subsection, a series of experiments are conducted to study the effect of hyperparameters.

#### 5.9.1. Subgraph size K

DSLAD is executed on six benchmark datasets with subgraph size $K$ within [2,10]. It is seen from Figure 4 that AUC grows until the peak and then drops with subgraph size $K$ increasing. DSLAD achieves the AUC peak at $K = 5$ on Citeseer and Flickr, at $K = 7$ on Pubmed, and $K = 4$ on the rest datasets. These results show that too small subgraph contains insufficient information, restricting anomaly detection; Too large subgraph contains tedious information, which would hurt our model; Applicable subgraph size guarantee DSLAD in best performance.

#### 5.9.2. Effect of hyperparameter $\alpha$ and $\lambda$

Besides hyperparameter subgraph size $K$, we also discuss $\alpha$ and $\lambda$.

To explore the effect of hyperparameter $\alpha$, we select its value from {0.2, 0.4, 0.6, 0.8}. As illustrated in Figure 5, when $\alpha = 0.6$, DSLAD has the best performance on Citeseer, BlogCatalog, and Flickr. When $\alpha = 0.8$, DSLAD has the best performance on Cora, Pubmed, and ACM. This shows that no matter any datasets, the context anomaly score has a greater impact than the reconstruction anomaly score, supporting the assertion that the context score affects more.

To explore the effect of hyperparameter $\lambda$, we select its value from {0.5, 1.0, 1.5, 2.0, 2.5, 3.0}. As illustrated in Figure 5, DSLAD has the best performance when $\lambda = 3.0$, $\lambda = 3.0$, $\lambda = 2.0$, $\lambda = 1.0$, $\lambda = 2.5$, $\lambda = 1.5$ on Cora, Citeseer, Pubmed, BlogCatalog, Flickr and ACM respectively. A suitable setting of the hyperparameter $\lambda$ could prompt hidden space mapping and provide significant benefits for anomaly detection.

## 6. Conclusion

In this paper, a novel framework called DSLAD is proposed for graph anomaly detection. DSLAD is composed of four modules: discrimination pair sampling, GNN-based embedding, anomaly discrimination and contrastive representation learning. Both contrastive learning and generative learning are employed to discriminate anomaly. They complement one another and improve the effectiveness. The contrastive representation learning, greatly alleviating the semantic mixture and imbalance problem, generates a more balanced semantic space and facilitates node embedding.

By decoupling anomaly discrimination and contrastive representation learning, the performance of DSLAD is undoubtedly improved. In the future, we will expore a unified representation learning framework for anomaly detection and node classification.

## 7. ACKNOWLEDGEMENT

The research is supported by the Key-Area Research and Development Program of Guangdong Province (2020B0101 65003, 2020B0101090005), the National Natural Science Foundation of China (62176269), the Guangdong Basic and Applied Basic Research Foundation (2019A1515011043), the Innovative Research Foundation of Ship General Performance (25622112), and the National Natural Science Foundation of China and Guangdong Provincial Joint Fund (U1911202).

## CRediT authorship contribution statement

**YanMing Hu:** Conceptualization, Methodology, Software, Validation, Formal analysis, Investigation, Writing - Original Draft, Visualization, Formal analysis. **Chuan Chen:** Conceptualization, Methodology, Writing - Original Draft, Supervision. **BoWen Deng:** Validation, Writing - Original Draft. **YuJing Lai:** Writing - Original Draft. **Hao Lin:** Resources, Project administration. **ZiBin Zheng:** Resources, Project administration. **Jing Bian:** Supervision.